\documentclass[10pt,twocolumn,letterpaper]{article}

\usepackage{cvpr}
\usepackage{times}
\usepackage{epsfig}
\usepackage{graphicx}
\usepackage{amsmath}
\usepackage{amssymb}
\usepackage{booktabs} 
\usepackage{algorithm}
\usepackage{subfigure}
\usepackage[noend]{algpseudocode}
\usepackage[pagebackref=true,breaklinks=true,letterpaper=true,colorlinks,bookmarks=false]{hyperref}

\cvprfinalcopy

\ifcvprfinal\fi
\begin{document}

\title{Evaluating Generalization Ability of Convolutional Neural Networks and Capsule Networks for Image Classification via Top-2 Classification}

\author{Hao Ren\\
School of Computer Science\\
Fudan University\\
Shanghai, China\\
{\tt\small hren17@fudan.edu.cn}
\and
Jianlin Su\\
School of Mathematics\\
Sun Yat-sen University\\
Guangzhou, China\\
{\tt\small bojone@spaces.ac.cn}
\and
Hong Lu\\
School of Computer Science\\
Fudan University\\
Shanghai, China\\
{\tt\small honglu@fudan.edu.cn}
}

\maketitle

\begin{abstract}
Image classification is a challenging problem which aims to identify the category of object in the image. In recent years, deep Convolutional Neural Networks (CNNs) have been applied to handle this task, and impressive improvement has been achieved. However, some research showed the output of CNNs can be easily altered by adding relatively small perturbations to the input image, such as modifying few pixels. Recently, Capsule Networks (CapsNets) are proposed, which can help eliminating this limitation. Experiments on MNIST dataset revealed that capsules can better characterize the features of object than CNNs. But it's hard to find a suitable quantitative method to compare the generalization ability of CNNs and CapsNets. In this paper, we propose a new image classification task called Top-2 classification to evaluate the generalization ability of CNNs and CapsNets. The models are trained on single label image samples same as the traditional image classification task. But in the test stage, we randomly concatenate two test image samples which contain different labels, and then use the trained models to predict the top-2 labels on the unseen newly-created two label image samples. This task can provide us precise quantitative results to compare the generalization ability of CNNs and CapsNets. Back to the CapsNet, because it uses Full Connectivity (FC) mechanism among all capsules, it requires many parameters. To reduce the number of parameters, we introduce the Parameter-Sharing (PS) mechanism between capsules. Experiments on five widely used benchmark image datasets demonstrate the method significantly reduces the number of parameters, without losing the effectiveness of extracting features. Further, on the Top-2 classification task, the proposed PS CapsNets obtain impressive higher accuracy compared to the traditional CNNs and FC CapsNets by a large margin. It shows the PS CapsNets have a much better generalization ability than the previous CNNs and FC CapsNets. In addition, to better understand the behavior of CapsNets, we develop a novel visualization approach called Probability-guided Activation Mapping (ProbAM) to obtain the focused parts in the images. This method can provide us qualitative results to compare the generalization ability of CNNs and CapsNets.
\end{abstract}

\section{Introduction}

Recent years we have seen increased research attention being directed towards image classification, with the rapid development of deep Convolutional Neural Networks (CNNs). Specifically, after Krizhevsky \etal \cite{krizhevsky2012imagenet} proposed AlexNet, which got the first place of 2012 ImageNet Large Scale Visual Recognition Challenge (ILSVRC) \cite{russakovsky2015imagenet}, CNNs have been applied to many computer vision tasks \cite{wang2013learning,ren2015faster,pinheiro2015learning}. The CNN based methods achieved better performance with deeper structure. Nevertheless, with the growth of layers, training procedure encounters the problem of gradient vanishing.

To conquer this weakness, Residual Network (ResNet) \cite{he2016deep} was proposed. It introduced shortcut connection between residual blocks, and achieved promising performance on 2015 ILSVRC. Thereafter, various variants of ResNet have been developed \cite{huang2016deep,xie2017aggregated,huang2017densely}, making the performance further improved. However, Su \etal \cite{su2017one} pointed out the output of CNNs is sensitive to tiny perturbation on the input image, and the network would be defrauded by changing few pixels.

On the other hand, Hinton \etal \cite{hinton2011transforming} presented capsule, which is a small group of neurons and can be regarded as a vector. The activities of neurons are used to represent various properties of an entity. Sabour \etal \cite{sabour2017dynamic} applied this concept to neural network firstly. It developed a Fully Connected (FC) capsule layer. And a novel routing algorithm called dynamic routing was adopted to select active capsules. The CapsNet was experimented on MNIST dataset \cite{lecun1998gradient}, attained state-of-the-art performance. Further, experiments of CapsNet showed capsules could learn a more robust representation than traditional CNNs. However, there are many parameters between FC capsule layers. This limitation leads to the impossibility of classifying large image datasets, such as ImageNet.

To reduce the number of parameters between capsule layers and more effectively select the active capsules, Hinton \etal \cite{e2018matrix} proposed matrix capsule to replace vector capsule. EM routing algorithm was introduced to obtain the agreement between capsules. Experiments on smallNORB dataset \cite{lecun2004learning} showed the proposed approach achieved state-of-the-art performance. In addition, it also revealed that capsules have the ability of withstanding white box adversarial attack than the baseline CNN. Despite it had fewer parameters than the vector CapsNet, it still required more parameters than CNN. Meanwhile, as it used matrix capsule to replace the vector capsule, it lost the ability of using capsule's length as activation, and had to allocate a scalar for each capsule as its activation.

Both the works of Sabour \etal \cite{sabour2017dynamic} and Hinton \etal \cite{e2018matrix} only present the simple qualitative results about CapsNets, have not present a detailed and intuitive method to compare the generalization ability of CNNs and CapsNets. So in this paper, we propose a new image classification task called Top-2 classification to evaluate the generalization ability of CNNs and CapsNets. The task can provide detailed and precise quantitative results to compare the performance about these models. And we also develop a novel visualization approach called Probability-guided Activation Mapping (ProbAM) to provide intuitive qualitative results. To further reduce the number of parameters about CapsNet, we introduce a new capsule layer with parameter-sharing (abbreviated as PS capsule layer). We conduct a series of experiments on the Top-2 classification task. Experimental results demonstrate that our proposed PS CapsNets achieve much better generalization ability than corresponding CNNs and FC CapsNets on five benchmark image datasets. The main contributions of this paper are as follows:
\begin{itemize}
	\item We propose a new image classification task called Top-2 classification to better evaluate the generalization ability of CNNs and CapsNets.
	\item We develop a novel visualization approach called Probability-guided Activation Mapping (ProbAM) to obtain the focused parts in the images. This method can help us better understand the behavior of CapsNets.
	\item We introduce the Parameter-Sharing (PS) mechanism between capsules. This method significantly reduces the number of parameters.
	\item We conduct a series of experiments on the Top-2 classification task, the proposed PS CapsNets obtain impressive higher accuracy compared to the traditional CNNs and FC CapsNets by a large margin on five widely used benchmark image datasets.
\end{itemize}

\section{Our approaches}

It is natural to ask a follow-up question: how do we confirm whether a model has good generalization ability in the image classification task? The common solution is predicting the categories of objects in the unseen images. The model is trained on single label image samples, and then tested on the unseen image samples which also contain one label for each sample. The model just need to predict one category for each image. We think it does not seem to be very convincing for evaluating the generalization ability of a model.

So we design a more difficult and complicated task to confirm the generalization ability of a model, we call it Top-2 classification. To handle this task, we propose a new capsule layer, which called PS capsule layer. Subsequently, a novel visualization approach named ProbAM is developed to help us better understand the behavior of CapsNets. We elaborate our Top-2 classification task, PS capsule layer and ProbAM approach in detail as below.

\subsection{Top-2 classification}
\label{task}

Formally, suppose $\mathbf{I}=\{I_1, I_2, \dots, I_m \}$ to represent the image space of $m$ training samples, $\mathbf{X}=\{x_1, x_2, \dots, x_n \}$ to represent the image space of $n$ testing samples, $\mathbf{Y}=\{y_1, y_2, \dots, y_q \}$ to represent the label space of $q$ possible labels. Each training or testing sample is an image of shape $(c,h,w)$, $c$ means channel, $h$ means height, $w$ means width, and $\mathbf{I} \cap \mathbf{X}=\emptyset$.

The training dataset is defined as 
    \begin{align}
	    D_{train}=\big \{(I^i, y^i)\big \}, \quad 1 \leqslant i \leqslant m
    \end{align}
    where $I^i \in \mathbf{I}, \quad y^i \in \mathbf{Y}$, $\quad y^i$ is the corresponding label of $I^i$.
        
For the traditional image classification task, the testing dataset is defined as 
    \begin{align}
	    D_{test}=\big \{(x^j, y^j)\big \}, \quad 1 \leqslant j \leqslant n
    \end{align}
    where $x^j \in \mathbf{X}, \quad y^j \in \mathbf{Y}$, $\quad y^j$ is the corresponding label of $x^j$. The aim of this task is learning a model $f(\cdot)$ to predict the true label $y^j$ for the image $x^j$.
    
For the Top-2 image classification task, the testing dataset is defined as 
    \begin{align}
    	D^{'}_{test}=\Big \{ \big (x^k \odot x^l, \{ y^k, y^l \} \big )\Big \}, \quad 1 \leqslant k, l \leqslant n
    \end{align}
    where $x^k, x^l \in \mathbf{X}, \quad y^k, y^l \in \mathbf{Y}$, $\quad y^k$ is the corresponding label of $x^k$, $\quad y^l$ is the corresponding label of $x^l$, $\quad k \neq l$ and $y^k \neq y^l$. The $\odot$ means concatenation operation, $x^k \odot x^l$ is an image of shape $(c,h,2w)$. The aim of this task is learning a model $f^{'}(\cdot)$ to predict the true label $y^k$ and $y^l$ for the image $x^k \odot x^l$.

Both the model $f(\cdot)$ and $f^{'}(\cdot)$ are trained on the same training dataset $D_{train}$. In the image classification domain, the universal output of models is a vector of $q$ dimension. Each component of this vector is used to represent the probability of corresponding label. So we can represent the model $f(\cdot)$ and $f^{'}(\cdot)$ as $H(\cdot)$ in a uniform way. For the traditional image classification task, the model $H(\cdot)$ outputs the corresponding label with highest probability. For the Top-2 image classification task, the model $H(\cdot)$ outputs the corresponding labels with top-2 highest probabilities. 

From the definitions of these two tasks, it is obvious that the Top-2 image classification task is more difficult and complicated. The model $H(\cdot)$ is only trained on images with single label, but tested on the images with two labels. We think this task is more suitable than traditional image classification task to evaluate the generalization ability of models. We conduct experiments to confirm this in Section~\ref{experiments}.

\subsection{Parameter-Sharing capsule layer}

Suppose there are $N$ low-level capsules
    \begin{align}
	    U= \{ \boldsymbol{u}_1, \boldsymbol{u}_2, \boldsymbol{u}_i, \dots, \boldsymbol{u}_N \}, \quad 1 \leqslant i \leqslant N
    \end{align}
and $M$ high-level capsules
    \begin{align}
	    V= \{ \boldsymbol{v}_1, \boldsymbol{v}_2, \boldsymbol{v}_j, \dots, \boldsymbol{v}_M \}, \quad 1 \leqslant j \leqslant M
    \end{align}
where $\boldsymbol{u}_i$ and $\boldsymbol{v}_j$ are vectors, the dimension of them can be different. The goal of capsule layer is clustering the $N$ low-level capsules to $M$ high-level capsules. The capsule layer firstly receives low-level capsules, which can be regarded as low-level features, then uses transform matrixes to transform low-level capsules to target feature domain. Finally, the routing algorithm clusters the transformed capsules to high-level capsules. 

\begin{figure}
  \centering
  \includegraphics[width=0.9\linewidth]{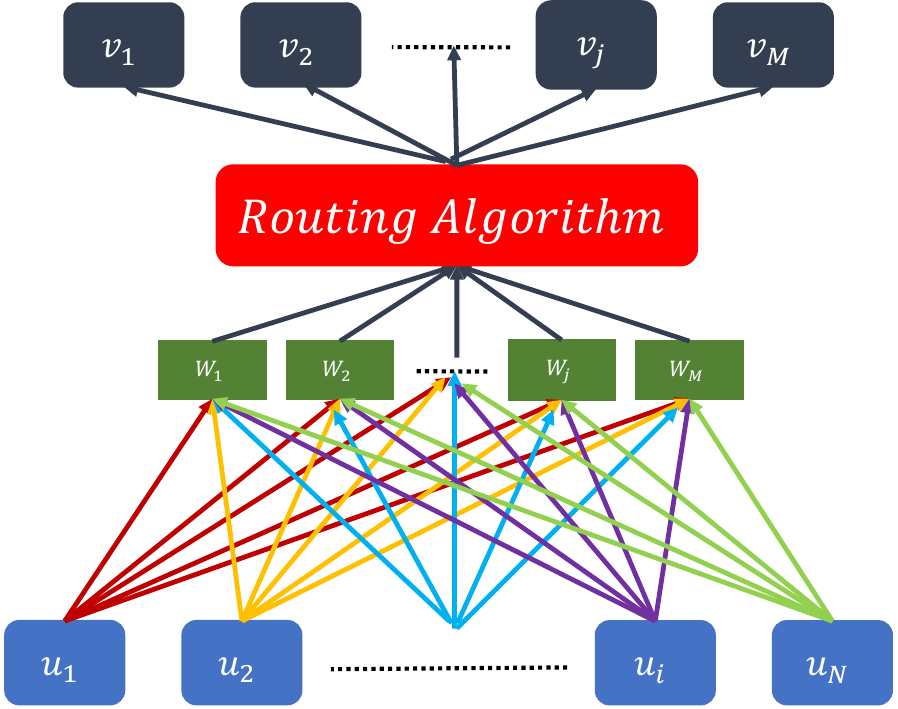}
  \caption{Parameter-Sharing capsule layer, where $M$ denotes the number of high-level capsules, $N$ denotes the number of low-level capsules, $\boldsymbol{u}_i$ denotes the low-level capsule, $\boldsymbol{v}_j$ denotes the high-level capsule, and $\boldsymbol{W}_j$ denotes the transform matrix.}
  \label{capsule_layer}
\end{figure}

The structure of PS capsule layer is shown in Fig.~\ref{capsule_layer}. The transform matrix is regarded as a feature selector for each high-level capsule, without considering the position information of low-level capsules. Unlike FC capsule layer, which uses exclusive transform matrix between each low-level capsule and high-level capsule, PS capsule layer shares the transform matrix between low-level capsules. Then the FC capsule layer needs $M \times N$ transform matrixes, but PS capsule layer needs only $M$ transform matrixes in this case. So it is obvious that PS capsule layer needs fewer parameters than FC capsule layer.

Routing algorithm is the core of capsule layer. As mentioned before, we need a method to cluster the low-level capsules and form the high-level capsules. The work by Ren and Lu \cite{ren2018compositional} presented a new routing algorithm named k-means routing recently. The work showed that k-means routing is more robust and stable than the dynamic routing, which is proposed by Sabour \etal \cite{sabour2017dynamic}. Both these two routing algorithms adopted a $squash$ function:
    \begin{align}
      squash(\boldsymbol{v}_j)= \frac{\Vert \boldsymbol{v}_j \Vert}{1 + \Vert \boldsymbol{v}_j \Vert^2} \boldsymbol{v}_j
    \end{align}
to make sure the length of capsule between 0 and 1. 

In our work, we decide using k-means routing as our routing algorithm to cluster the capsules. But we use the dot product value to measure the similarity between each low-level capsule and high-level capsule, because dot product has lower computational complexity than cosine similarity. This is different from k-means routing, which uses cosine similarity. As mentioned, since PS capsule layer shares the transform matrix between low-level capsules, we needn't $M \times N$ transform matrixes, the total number of transform matrixes is $M$. The modified routing procedure is summarized in Algorithm~\ref{routing}. 

\begin{algorithm}[H]
  \caption{Modified K-means Routing}
  \label{routing}
  \begin{algorithmic}[1]
    \Procedure {Routing}{$\boldsymbol{u}_i, r$}
    \State Initialize $\boldsymbol{v}_j \leftarrow \frac{1}{M}  \sum\limits_{i=1}^N \boldsymbol{W}_j \boldsymbol{u}_i$
	 \For {$r$ iterations}
	  \State $b_{ij} \leftarrow \frac{(\boldsymbol{W}_j \boldsymbol{u}_i) \cdot \boldsymbol{v}_j}{\Vert \boldsymbol{v}_j \Vert}$
	  \State $c_{ij} \leftarrow \mathop{softmax}\limits_j b_{ij}$
	  \State $\boldsymbol{v}_j \leftarrow \sum\limits_{i=1}^N c_{ij} \boldsymbol{W}_j \boldsymbol{u}_i$
	\EndFor
	\State \Return $squash(\boldsymbol{v}_j)$
	\EndProcedure
  \end{algorithmic}
\end{algorithm}

\begin{figure*}
  \centering
  \includegraphics[width=\linewidth]{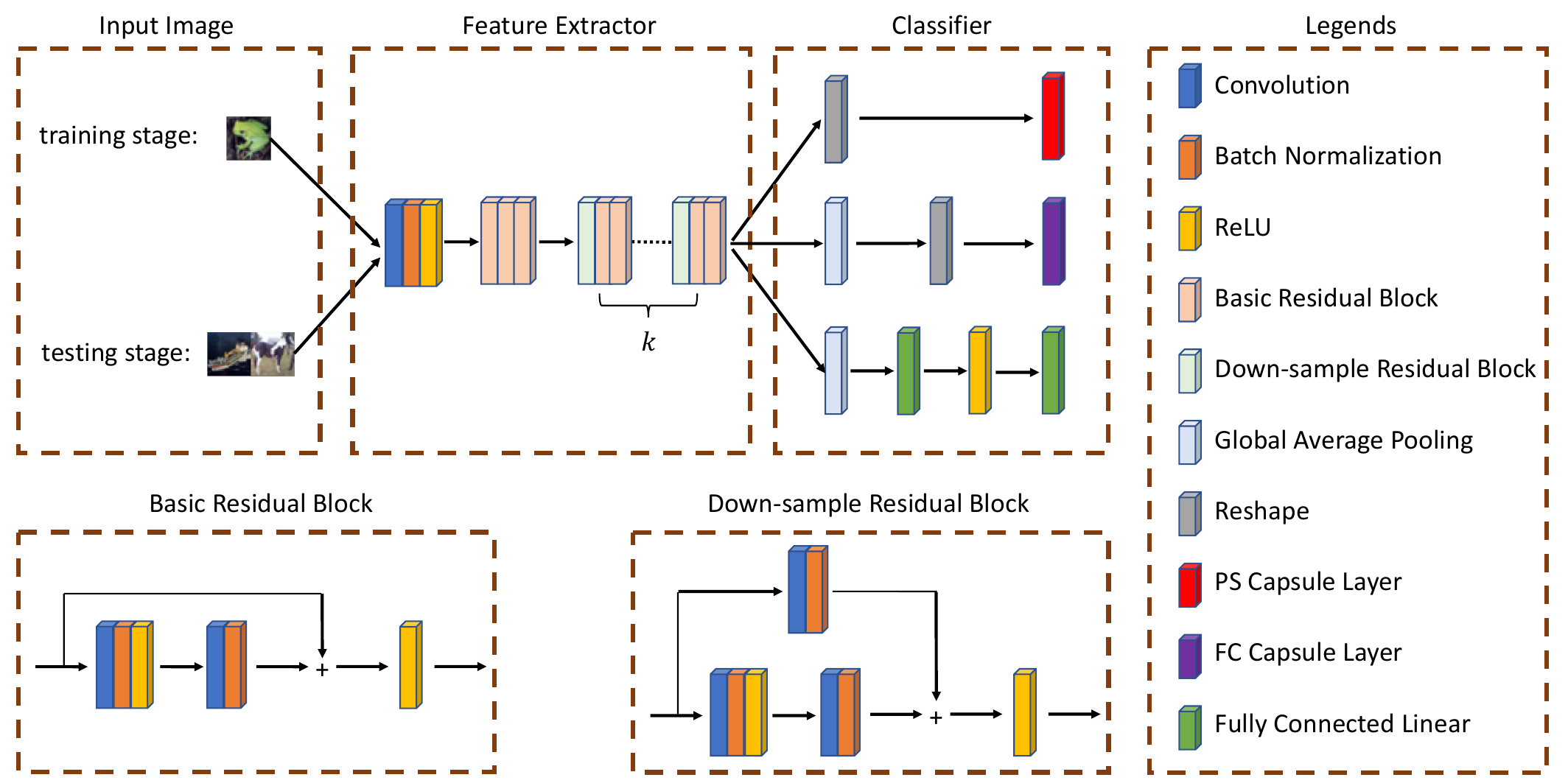}
  \caption{Model structure, where \emph{Reshape} denotes the operation that divide the input feature map into capsules along channel dimension. In the Classifier, from top to bottom, the structure is designed for PS Capsule based, FC Capsule based and CNN based respectively.}
  \label{model_structure}
\end{figure*}

\subsection{Model architecture}

In this section, we describe three models designed to handle the Top-2 image classification task. The three models are CNN based, FC Capsule based and PS Capsule based. The architectures of these models all can be divided into two parts: feature extractor and classifier. The feature extractor extracts features and feeds them into classifier. The classifier takes advantage of them to correctly classify the images. Because the aim of this work is to evaluate and compare the generalization ability of CNNs, FC CapsNets and our proposed PS CapsNets, the three models use exactly same feature extractor structure, the only difference is the classifier. The overall structures of these models are illustrated in Fig.~\ref{model_structure}. As the figure shows, the models are trained on single label images, and tested on two-label images. We elaborate the feature extractor and classifier in detail as below.

\subsubsection{Feature extractor}

As Fig.~\ref{model_structure} shows, the three models use the same feature extractor. The feature extractor contains traditional convolution layer, batch normalization layer \cite{ioffe2015batch} and ReLU activation function. It is similar to ResNet. We take the model designed for CIFAR10 dataset \cite{krizhevsky2009learning} as an example, it consists of seven basic residual blocks and two down-sample residual blocks, which means $k=2$. The kernel size of convolution layer is $3 \times 3$, except the top convolution layer of down-sample residual block, where the kernel size is $1 \times 1$. All the convolution layers use padding, without bias. Stride=2 on the first convolution layer and the top convolution layer of down-sample residual block. The output channels of the first convolution layer and the first three blocks are $(16,16)$, the next output channels of the remaining blocks are $(32,64)$. 

In our work, we conduct experiments on five benchmark image datasets. The models designed for other four datasets have the same structure, where $k=2$, except STL10 dataset \cite{coates2011analysis}, where $k=3$. The details of these datasets are summarized in Section~\ref{datasets}.
 
\begin{figure*}
  \centering
  \includegraphics[width=\linewidth]{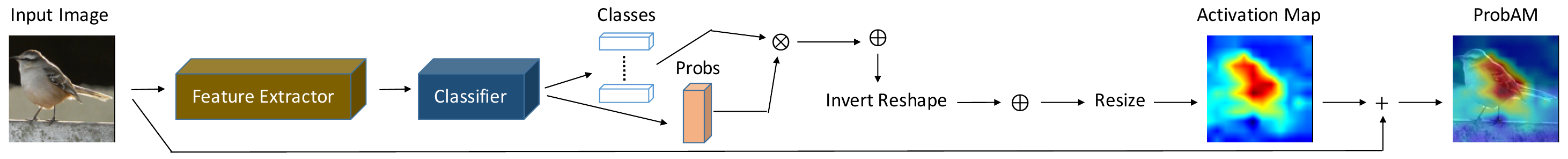}
  \caption{ProbAM procedure: Given an image as input, we forward propagate the image through the model to obtain high-level capsules (which is denoted by \emph{Classes}) and probabilities (which is denoted by \emph{Probs}, they are the set of $c_{ij}$, and $c_{ij}$ was defined in Algorithm \ref{routing}). Then we compute the element-wise product of them, sum up along high-level capsule dimension. This value is then recovered to last layer's feature map of Feature Extractor (where \emph{Invert Reshape} is the reverse process of \emph{Reshape}, which is defined in Fig.~\ref{model_structure}). Finally, we sum it up along channel dimension, and resize to the same size as input image to get the activation map. ProbAM is generated by adding activation map and input image.}
  \label{probam_procedure}
\end{figure*}

\subsubsection{Classifier}

We can see from Fig.~\ref{model_structure} that the classifier contains three different structures. From top to bottom, the structure is designed for PS Capsule based, FC Capsule based and CNN based models respectively. For the PS Capsule based model, the classifier first reshapes the feature map into several capsules, and then feeds them into the PS capsule layer. 

As mentioned in Section~\ref{task}, the training image sample's shape is $(c,h,w)$, the testing image sample's shape is $(c,h,2w)$. For the FC capsule layer and fully connected linear layer, the number of parameters relies on the input image's shape. The common solution is global average pooling \cite{lin2014network}. So the feature map is firstly downsampled to a fixed size $(4,4)$ by a global average pooling operation. Then the output feature map is reshaped into several capsules, and fed into the FC capsule layer. For CNN based model, we adopt the universal structure of fully connected linear layer.

The length of low-level capsules is 32, and the length of high-level capsules is 8. The output features of first fully connected linear layer is 256. We apply $sigmoid$ function to the output features for CNN based model, and use the length of output capsule to represent the probability of label for Capsule based models.

\subsection{Probability-guided activation mapping}

Class Activation Mapping (CAM, \cite{zhou2016learning}) is a sort of methods to visualize the feature map and highlight the areas where the network focuses on by given specific class. It can help us understand how the network works. Grad-CAM \cite{selvaraju2017grad} is the representative method among this sort of methods. But it only works for CNN based models, and is not suitable for Capsule based models. Furthermore, it needs computing gradients from backward and be given specific target class beforehand. 

In order to eliminate those limitations, we propose an original activation mapping method for Capsule based models. There is no need to give target class in advance, the method generates activation map from forward procedure, and needn't compute gradients.

The generation of activation map relies on the probabilities which are obtained from routing algorithm. That's why we call it Probability-guided Activation Mapping (ProbAM). As mentioned before, the routing algorithm aims to cluster low-level capsules, and form high-level capsules. So we can obtain the information about which low-level capsules are selected, and how much contribution they make to construct high-level capsules. After that, we map them back to the feature map, because the low-level capsules are reshaped from the feature map. The complete process is described in Fig.~\ref{probam_procedure}.

\section{Experiments}
\label{experiments}

In this section, we conduct extensive experiments on five benchmark image datasets for the Top-2 image classification task. We present the dataset, experimental settings and experimental results in the following sections. 

\subsection{Datasets}
\label{datasets}

Five widely applied benchmark image datasets are used on our experiments. These datasets are MNIST handwritten digits dataset  \cite{lecun1998gradient}, FashionMNIST dataset \cite{xiao2017fashion}, Street View House Numbers (SVHN) dataset \cite{netzer2011reading}, CIFAR10 dataset \cite{krizhevsky2009learning}, and STL10 dataset \cite{coates2011analysis}.

The models are trained on the single-label training images from those datasets, and first tested on the unseen single-label testing images. It provides us the classification accuracy of single-label images. After that, we generate two-label testing images from single-label testing images by randomly sampling two images which have different labels, and concatenate the two images along width dimension. We obtain the top-2 outputs' corresponding labels, and compare them with the ground truth to obtain the two-label image classification accuracy. Table~\ref{datasets_table} shows the summary of these five datasets. 

\begin{table}
  \centering
  \caption{Statistics of the benchmark image classification datasets, where SL means single-label image, TL means two-label image.}
  \label{datasets_table}
  \begin{tabular}{l|lll}
    \toprule
    Dataset      & \# Train & \# Test (SL) & \# Test (TL) \\
    \midrule 
    MNIST        & 60,000   & 10,000       & 9,002 \\
    FashionMNIST & 60,000   & 10,000       & 8,961 \\
    SVHN         & 73,257   & 26,032       & 22,900 \\ 
    CIFAR10      & 50,000   & 10,000       & 9,005 \\
    STL10        & 5,000    &  8,000       & 7,181 \\
    \bottomrule
  \end{tabular}
\end{table}

\begin{table*}
  \centering
  \caption{Quantitative test accuracy results of CNNs, FC CapsNets and PS CapsNets on the benchmark image classification datasets.}
  \label{table_accuracy_results}
  \begin{tabular}{l|lll|lll|lll}
    \toprule
    Dataset      & CNN-SA           & FC-SA            & PS-SA            & CNN-TA  & FC-TA   & PS-TA            & CNN-TCA & FC-TCA  & PS-TCA \\
    \midrule
    MNIST        & 99.65\%          & 99.64\%          & \textbf{99.66\%} & 62.74\% & 94.35\% & \textbf{99.41\%} & 11.95\% & 73.12\% & \textbf{99.41\%} \\
    FashionMNIST & \textbf{94.80\%} & 94.66\%          & 93.92\%          & 33.44\% & 44.88\% & \textbf{88.34\%} & 7.59\%  & 3.47\%  & \textbf{87.26\%} \\
    SVHN         & \textbf{96.48\%} & 96.28\%          & 96.24\%          & 36.46\% & 20.36\% & \textbf{84.54\%} & 9.64\%  & 2.34\%  & \textbf{84.10\%} \\
    CIFAR10      & 89.83\%          & \textbf{90.01\%} & 88.88\%          & 40.16\% & 67.15\% & \textbf{77.80\%} & 8.00\%  & 27.54\% & \textbf{76.90\%} \\
    STL10        & 76.15\%          & \textbf{77.08\%} & 75.36\%          & 25.18\% & 32.74\% & \textbf{54.25\%} & 6.36\%  & 5.51\%  & \textbf{52.79\%} \\
    \bottomrule
  \end{tabular}
\end{table*}

\subsection{Experimental settings}

We implemented our models with PyTorch library \cite{collobert2011torch7}. All the experiments are performed on a single NVIDIA GeForce GTX 1070 GPU. Margin loss \cite{sabour2017dynamic} is used to compute our models' loss:
    \begin{align}
      L=&\frac{1}{M} \sum_{j=1}^M \Big( T_j \max(0, 0.9-\Vert \boldsymbol{v}_j \Vert)^2 \\ \notag 
        & + 0.5(1-T_j) \max(0, \Vert \boldsymbol{v}_j \Vert-0.1)^2 \Big)    	
    \end{align}
where $T_j = 1$ iff an image of class $j$ is present, it is optimized through ADAM scheme \cite{kingma2015adam} with default learning rate and momentum. We set batch size to 64 and train the models with 100 epochs. The number of routing iterations is fixed to 3. The experiments are performed with data augmentation, such as normalization, random crop and random horizontal flip. The code of our work is available on \url{https://github.com/leftthomas/PSCapsNet}.

We record the following performance indexes of the experiments: single-label image classification accuracy (SA), two-label image classification accuracy (TA), two-label image classification accuracy with confidence level $\geqslant 50\%$ (TCA).

\subsection{Quantitative results}
\label{quantitative_results}

To validate the effectiveness of PS CapsNets, and compare the generalization ability with CNNs and FC CapsNets, we demonstrate a series of experiments on the five benchmark image classification datasets. The precise quantitative results about SA, TA and TCA are presented. 

We take the experiments on FashionMNIST dataset as an example, test accuracy is plotted after each training step, as shown in Fig.~\ref{experiment_fashionmnist} (more experimental results can be found in the supplementary material). We can observe that these three models obtain comparable SA. But when we test the two-label images, the accuracy drops dramatically for CNN and FC CapsNet, especially when we consider the confidence level. CNN and FC CapsNet are almost impossible to classify the two-label images correctly. On the other hand, we find PS CapsNet works very well, the accuracy only drops a little. What's more, TCA is close to TA for PS CapsNet. It means that PS CapsNet can recognize the objects with high confidence level.  All the results of five datasets show this characteristics. The overall quantitative results are shown in Table~\ref{table_accuracy_results}, better results are marked with bold. 
 
From the results, we can find CNNs and FC CapsNets works fine on single-label images, but they works awfully on two-label images. We even could say they are totally failed on some cases, \eg STL10 dataset. Meanwhile, our proposed PS CapsNets maintain good performance.  It can be concluded that our proposed PS CapsNets have a much better generalization ability than the traditional CNNs and FC CapsNets.

We also compare the number of parameters about these three models. The results are summarized in Table~\ref{parameters_table}, the minimum number of parameters are marked with bold. We observe that our PS CapsNets require fewer parameters than CNNs and FC CapsNets on all these five datasets. For example, the number of parameters on CIFAR10 dataset used in our PS CapsNet is $\sim 0.27M$, but the CNN needs more than $0.53M$ parameters, which is nearly 2 times than ours. The FC CapsNet requires $\sim 0.35M$ parameters, which is around 1.3 times than ours.

Take into account of test accuracy and model parameters, we can draw a conclusion that on the Top-2 image classification task, PS CapsNets achieve much better performance with fewer parameters compared with the CNNs and FC CapsNets. It is obvious that our PS CapsNets has huge advantage in practical application, and the results of this work can be a benchmark for the future works on the Top-2 image classification task. 

\begin{figure}
  \centering
  \includegraphics[width=\linewidth]{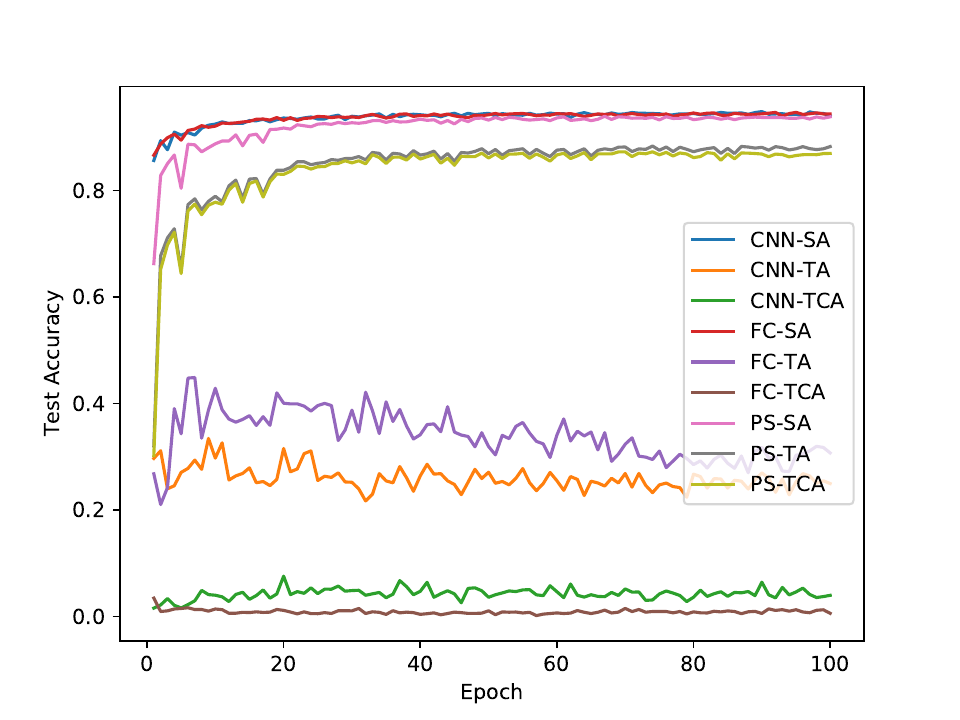}
  \caption{Test accuracy results of CNN, FC CapsNet and PS CapsNet after each training step on FashionMNIST dataset.}
  \label{experiment_fashionmnist}
\end{figure}

\begin{table}
  \centering
  \caption{Model parameters of CNNs, FC CapsNets and PS CapsNets on the benchmark image classification datasets.}
  \label{parameters_table}
  \begin{tabular}{l|lll}
    \toprule
    Dataset      & CNN     & FC      & PS \\
    \midrule 
    MNIST        & 536,506 & 353,456 & \textbf{274,096} \\
    FashionMNIST & 536,506 & 353,456 & \textbf{274,096} \\
    SVHN         & 536,794 & 353,744 & \textbf{274,384} \\ 
    CIFAR10      & 536,794 & 353,744 & \textbf{274,384} \\
    STL10        & 762,970 & 579,920 & \textbf{500,560} \\
    \bottomrule
  \end{tabular}
\end{table}

\begin{figure}
  \centering
  \includegraphics[width=\linewidth]{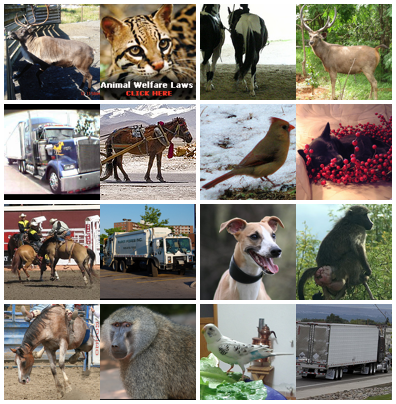}
  \caption{The original two-label images sampled from STL10. There are eight two-label images, which are divided by the white line.}
  \label{original_two_label_images}
\end{figure}

\begin{figure*} 
  \centering 
  \subfigure[CNN]{ 
    \includegraphics[width=0.32 \linewidth]{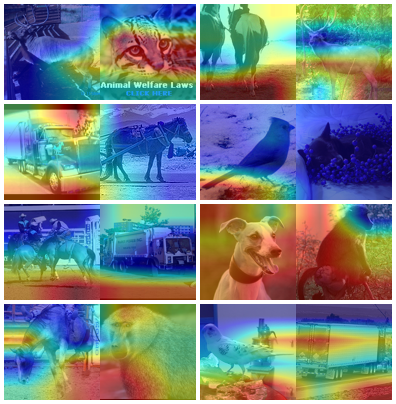} 
  } 
  \subfigure[FC CapsNet]{ 
    \includegraphics[width=0.32 \linewidth]{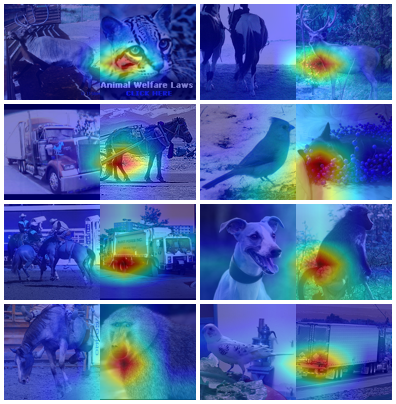} 
  }
  \subfigure[PS CapsNet]{ 
    \includegraphics[width=0.32 \linewidth]{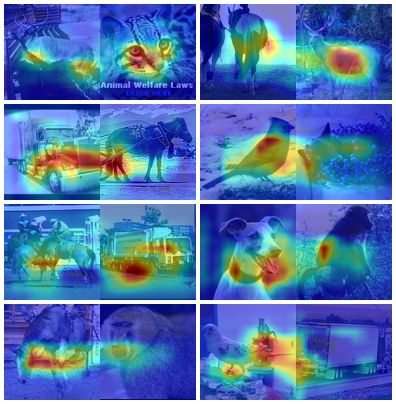}
    \label{vis_stl10_two_label_features_c} 
  }    
  \caption{The visualized last layer's feature map results of two-label images. There are eight two-label images, which are divided by the white line.} 
  \label{vis_stl10_two_label_features}
\end{figure*}

\begin{figure*}
  \centering
  \subfigure[CNN]{ 
    \includegraphics[width=0.32 \linewidth]{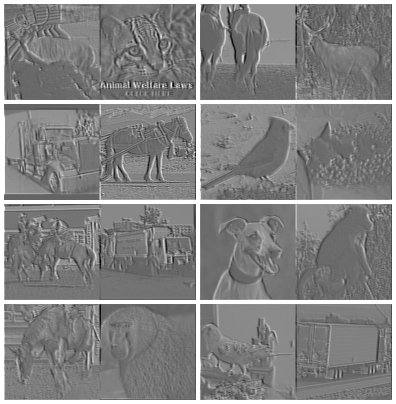} 
  } 
  \subfigure[FC CapsNet]{ 
    \includegraphics[width=0.32 \linewidth]{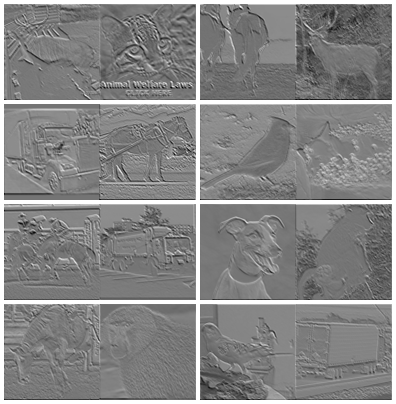} 
  }
  \subfigure[PS CapsNet]{ 
    \includegraphics[width=0.32 \linewidth]{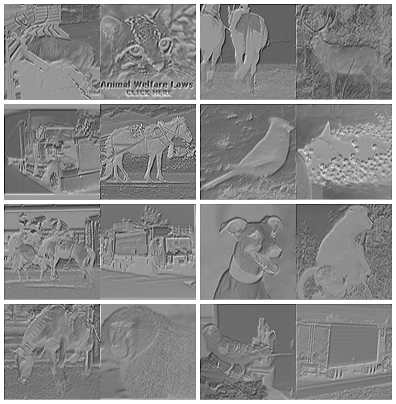} 
  } 
  \caption{The visualized first convolution layer's feature map results of two-label images. There are eight two-label images, which are divided by the white line.}
  \label{vis_stl10_two_label_conv1}
\end{figure*}

\subsection{Qualitative results}

To better understand how PS CapsNet works and why it works well, we use ProbAM to visualize the last layer's feature map of Feature Extractor, and explain why it focuses on those specific areas. We want to know how the PS CapsNets learn specific knowledge step by step, so we also visualize the first convolution layer to figure out the relationship between input image, lower layer and upper layer. The visualization of first convolution layer could be described by summing up the feature map along channel dimension. 

Meanwhile, to compare with the CNNs and FC CapsNets, we also use the proposed ProbAM method to visualize the feature map of FC CapsNets, and use the Grad-CAM to visualize the feature map of CNNs. Both these two visualization methods can provide us the focused areas in the images by the models. It is intuitive for human to understand the behavior of these models. 

Here we use the models experimented on STL10 dataset as an example. Fig.~\ref{original_two_label_images} shows the original two-label images. We randomly sample 8 two-label images, and feed them into the trained models to obtain the visualized results. Fig.~\ref{vis_stl10_two_label_features} shows the visualized last layer's feature map of Feature Extractor. And Fig.~\ref{vis_stl10_two_label_conv1} shows the visualized first convolution layer's feature map. More visualized results can be found in the supplementary material.

We can first observe from Fig.~\ref{vis_stl10_two_label_features} that PS CapsNet focuses on the objects very well in most cases, and it can distinguish objects and background. As it places great emphasis on the objects rather than the background, it generates high-level semantic information about those areas instead of other areas. That explains why it gives the correct labels of objects in the images. 

But when we look into the results of CNN, in most cases, the model can not distinguish objects and background. So the informations it provided are not enough to classify the objects correctly. It gives us the reason why CNNs work terribly on the Top-2 image classification task. Then let us consider the results of FC CapsNets, it is easy to find that FC CapsNets only focus on single object in each two-label images. The FC CapsNets even can not realize the existence of another object. That's the reason why it gets poor performance on the Top-2 image classification task.

Furthermore, we find our PS CapsNets still could generate great focal areas on the two objects, even though it is trained on single label images. That's the reason why our PS CapsNets achieve favorable performance on the two-label images. It also proves our PS CapsNets have good generalization ability. The qualitative results also corroborate the conclusion in Section~\ref{quantitative_results}.

If we look into the results carefully, we can find that our PS CapsNet haven't focus on the objects well for a few cases, such as the two-label image on the right bottom of Fig.~\ref{vis_stl10_two_label_features_c}. This image contains two objects: bird and truck. The truck is only focused by PS CapsNet a little. How can we interpret this? Let's take a look on Table~\ref{table_accuracy_results}, the model gets 75.36\% test accuracy on the single label images (PS-SA). So we know the model haven't learned enough, that's why the model can not focus perfectly for a few cases. It gives us a visualized results to help us analyse which class have been learnt enough or not.

We also find the relationship between input image, lower layer and upper layer from the results of Fig.~\ref{vis_stl10_two_label_conv1}. The visualized results of first convolution layer show the outline of objects in the images, and it removes some background noise. The three models all show this characteristic. This leads to a conclusion that the first few layers is in charge of learning overall information and eliminating noise, the following layers learn to extract specific features of objects. 

From the view of the principle of human vision system \cite{kandel2000principles}, when we recognize an object, we first remove the background noise, put our attention on this object, then we use specific features to decide what this object is. CNNs, FC CapsNets and our PS CapsNets all confirm exactly to this mechanism. The results also prove that our proposed ProbAM approach is an effective method to help us understand the behavior of CapsNets, and provide us an intuitive explanation.

\section{Conclusion}

In this paper, we proposed a new image classification task called Top-2 classification to evaluate the generalization ability of CNNs and CapsNets. A series of experiments have been conducted to validate the effectiveness of this task. The results show that this task is more suitable than the traditional image classification task to evaluate the generalization ability of CNNs and CapsNets. 

Then we introduced the Parameter-Sharing mechanism between capsules, and proposed PS capsule layer. This new capsule layer requires fewer parameters than the traditional FC capsule layer. We designed a simple model named PS CapsNet to test the performance of this method. Extensive experiments demonstrated PS CapsNet achieved impressive higher accuracy with fewer parameters compared with CNNs and FC CapsNets on the Top-2 classification task. 

Finally, we developed a novel activation mapping approach named ProbAM to obtain the parts on which CapsNets focused, and gave our explanations to help understand the working mechanism of CapsNets. This approach can be a useful tool to help us design a model with better generalization ability. 

{\small
\bibliographystyle{ieee}
\bibliography{egbib}
}

\appendix

\begin{figure*}[h]
  \centering
  \subfigure[MNIST]{ 
    \includegraphics[width=0.48 \linewidth]{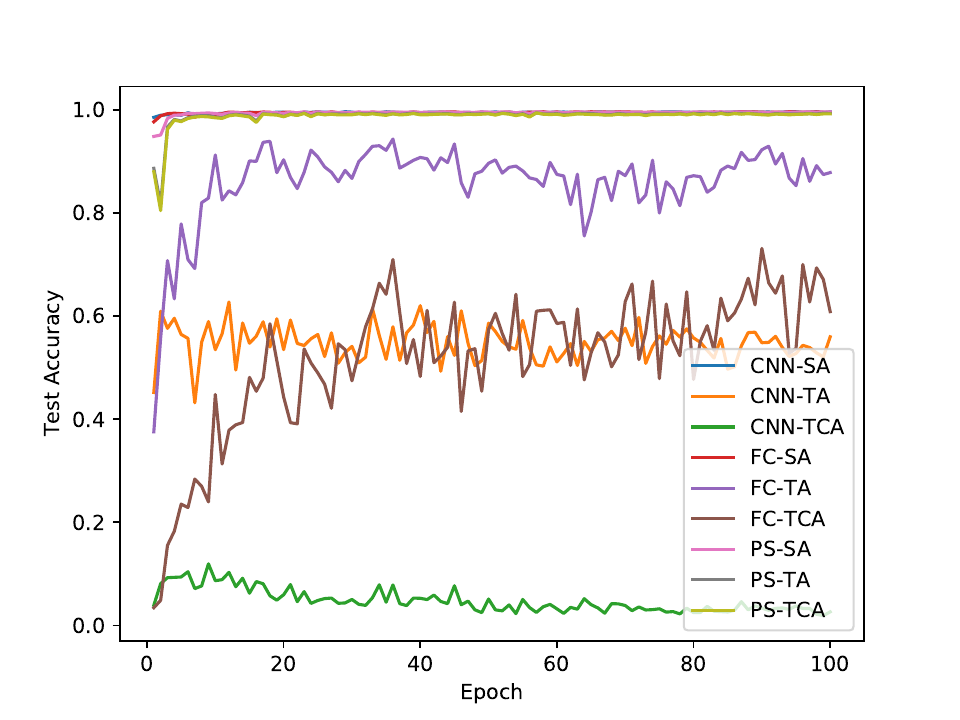} 
  } 
  \subfigure[SVHN]{ 
    \includegraphics[width=0.48 \linewidth]{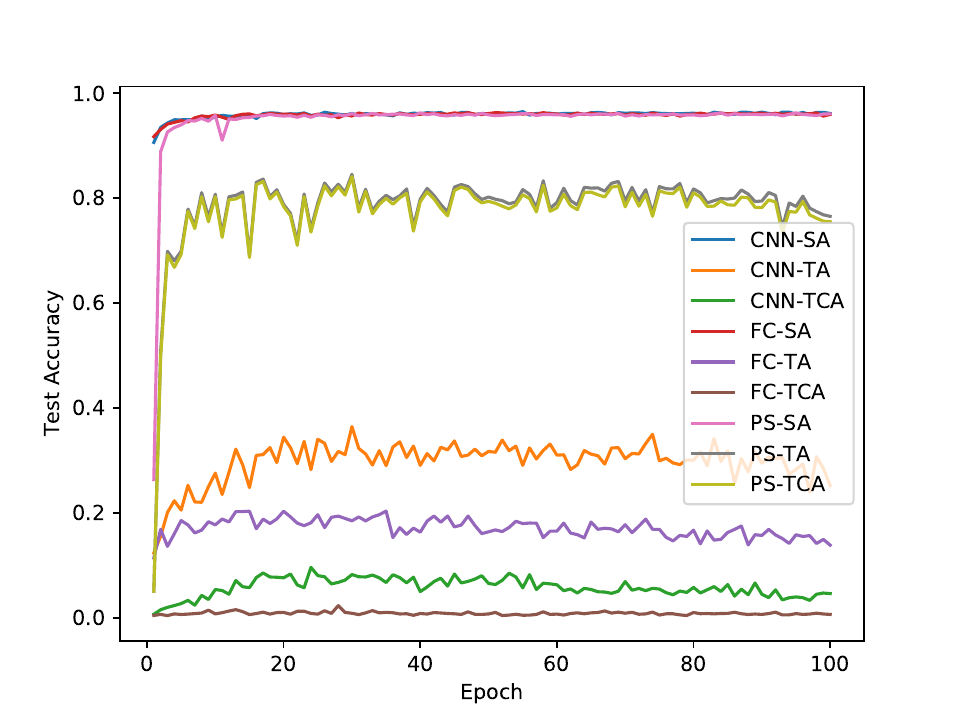} 
  }
  \subfigure[CIFAR10]{ 
    \includegraphics[width=0.48 \linewidth]{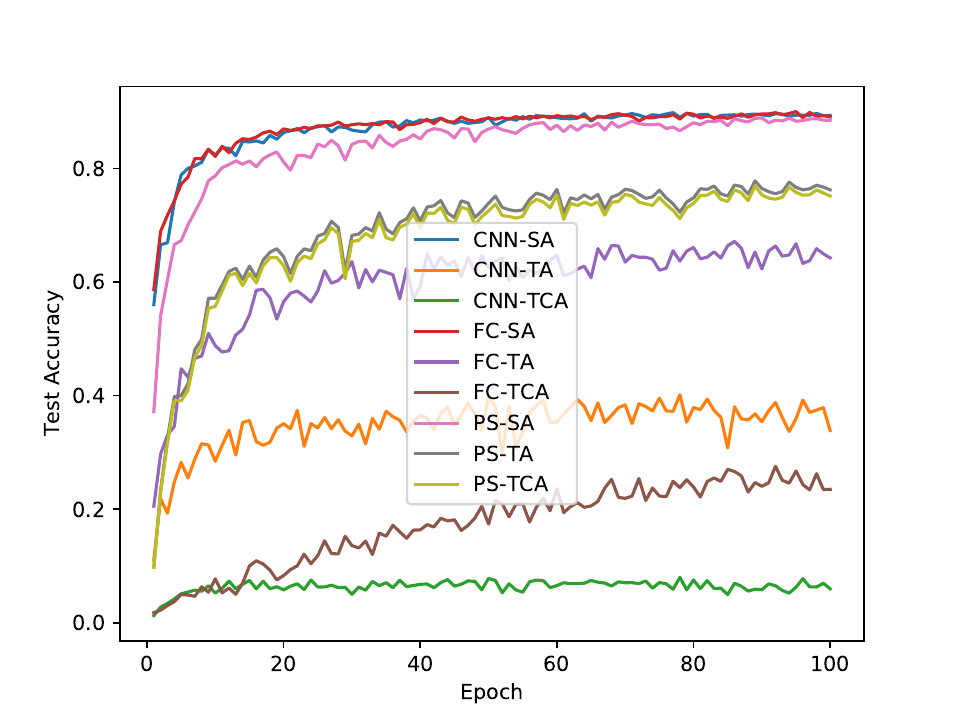} 
  } 
  \subfigure[STL10]{ 
    \includegraphics[width=0.48 \linewidth]{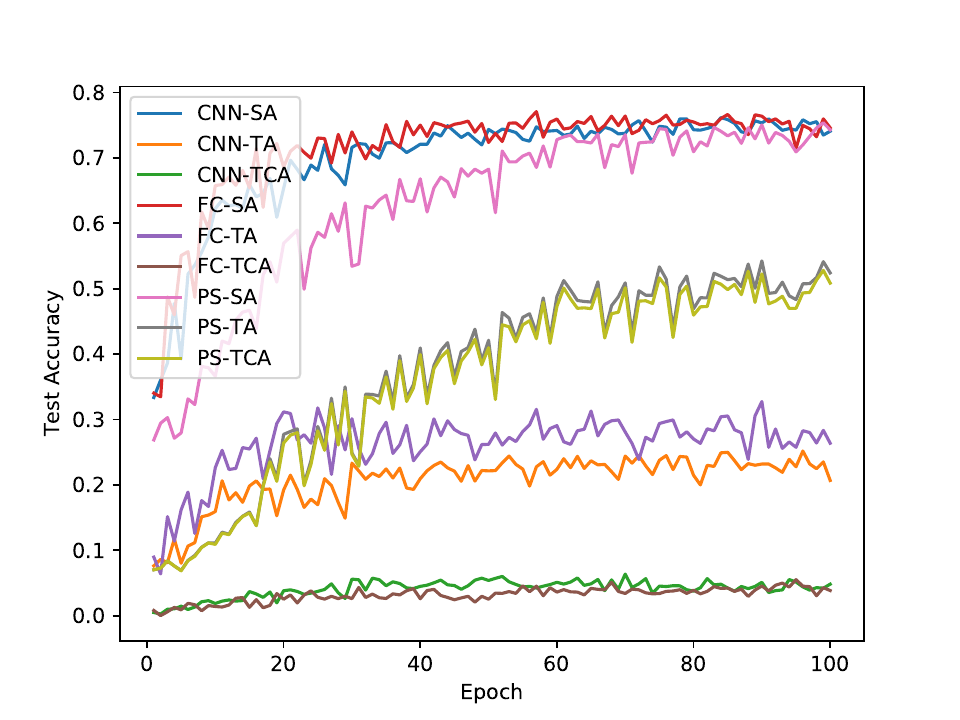} 
  }
  \caption{Test accuracy results of CNNs, FC CapsNets and PS CapsNets after each training step on the other four datasets.}
  \label{experiment_others}
\end{figure*}

\begin{figure*}
  \centering
  \subfigure[MNIST]{ 
    \includegraphics[width=0.18 \linewidth]{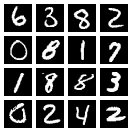} 
  } 
  \subfigure[FashionMNIST]{ 
    \includegraphics[width=0.18 \linewidth]{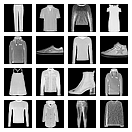} 
  }
  \subfigure[SVHN]{ 
    \includegraphics[width=0.18 \linewidth]{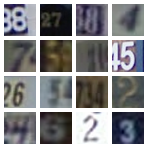} 
  } 
  \subfigure[CIFAR10]{ 
    \includegraphics[width=0.18 \linewidth]{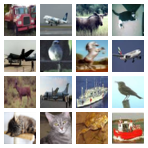} 
  }
  \subfigure[STL10]{ 
    \includegraphics[width=0.18 \linewidth]{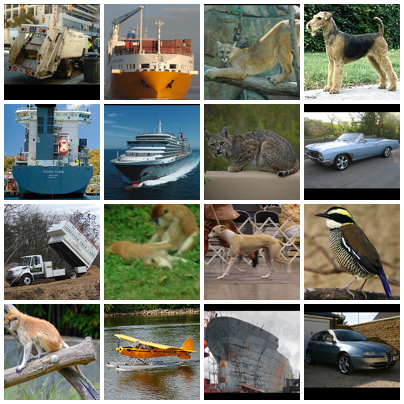} 
  }
  \caption{The original single-label images sampled from benchmark datasets. There are 16 single-label images, which are divided by the white line.}
  \label{original_single_label_images_others}
\end{figure*}

\begin{figure*}
  \centering
  \subfigure[MNIST]{ 
    \includegraphics[width=0.23 \linewidth]{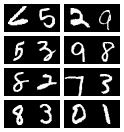} 
  } 
  \subfigure[FashionMNIST]{ 
    \includegraphics[width=0.23 \linewidth]{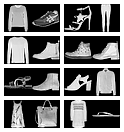} 
  }
  \subfigure[SVHN]{ 
    \includegraphics[width=0.23 \linewidth]{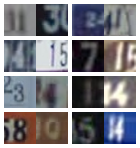} 
  } 
  \subfigure[CIFAR10]{ 
    \includegraphics[width=0.23 \linewidth]{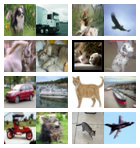} 
  }
  \caption{The original two-label images sampled from other datasets. There are eight two-label images, which are divided by the white line.}
  \label{original_two_label_images_others}
\end{figure*}

\begin{figure*}
  \centering
  \subfigure[CNN]{ 
    \includegraphics[width=0.32 \linewidth]{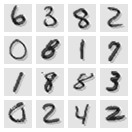} 
  } 
  \subfigure[FC CapsNet]{ 
    \includegraphics[width=0.32 \linewidth]{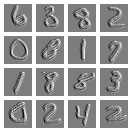} 
  }
  \subfigure[PS CapsNet]{ 
    \includegraphics[width=0.32 \linewidth]{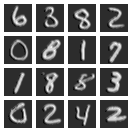} 
  } 
  \caption{The visualized first convolution layer's feature map results of single-label images on MNIST dataset. There are 16 single-label images, which are divided by the white line.}
  \label{vis_mnist_single_label_conv1}
\end{figure*}

\begin{figure*}
  \centering
  \subfigure[CNN]{ 
    \includegraphics[width=0.32 \linewidth]{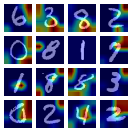} 
  } 
  \subfigure[FC CapsNet]{ 
    \includegraphics[width=0.32 \linewidth]{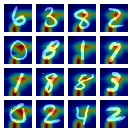} 
  }
  \subfigure[PS CapsNet]{ 
    \includegraphics[width=0.32 \linewidth]{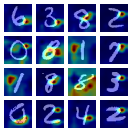} 
  } 
  \caption{The visualized last layer's feature map results of single-label images on MNIST dataset. There are 16 single-label images, which are divided by the white line.}
  \label{vis_mnist_single_label_features}
\end{figure*}

\begin{figure*}
  \centering
  \subfigure[CNN]{ 
    \includegraphics[width=0.32 \linewidth]{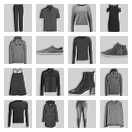} 
  } 
  \subfigure[FC CapsNet]{ 
    \includegraphics[width=0.32 \linewidth]{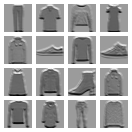} 
  }
  \subfigure[PS CapsNet]{ 
    \includegraphics[width=0.32 \linewidth]{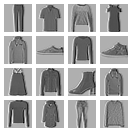} 
  } 
  \caption{The visualized first convolution layer's feature map results of single-label images on FashionMNIST dataset. There are 16 single-label images, which are divided by the white line.}
  \label{vis_fashionmnist_single_label_conv1}
\end{figure*}

\begin{figure*}
  \centering
  \subfigure[CNN]{ 
    \includegraphics[width=0.32 \linewidth]{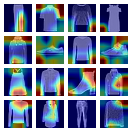} 
  } 
  \subfigure[FC CapsNet]{ 
    \includegraphics[width=0.32 \linewidth]{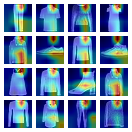} 
  }
  \subfigure[PS CapsNet]{ 
    \includegraphics[width=0.32 \linewidth]{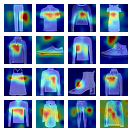} 
  }
  \caption{The visualized last layer's feature map results of single-label images on FashionMNIST dataset. There are 16 single-label images, which are divided by the white line.}
  \label{vis_fashionmnist_single_label_features}
\end{figure*}

\begin{figure*}
  \centering
  \subfigure[CNN]{ 
    \includegraphics[width=0.32 \linewidth]{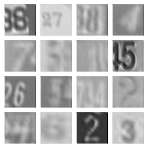} 
  } 
  \subfigure[FC CapsNet]{ 
    \includegraphics[width=0.32 \linewidth]{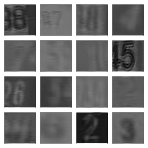} 
  }
  \subfigure[PS CapsNet]{ 
    \includegraphics[width=0.32 \linewidth]{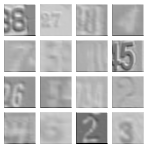} 
  }
  \caption{The visualized first convolution layer's feature map results of single-label images on SVHN dataset. There are 16 single-label images, which are divided by the white line.}
  \label{vis_svhn_single_label_conv1}
\end{figure*}

\begin{figure*}
  \centering
  \subfigure[CNN]{ 
    \includegraphics[width=0.32 \linewidth]{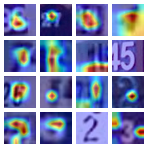} 
  } 
  \subfigure[FC CapsNet]{ 
    \includegraphics[width=0.32 \linewidth]{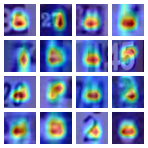} 
  }
  \subfigure[PS CapsNet]{ 
    \includegraphics[width=0.32 \linewidth]{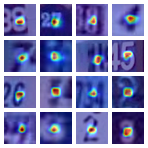} 
  }
  \caption{The visualized last layer's feature map results of single-label images on SVHN dataset. There are 16 single-label images, which are divided by the white line.}
  \label{vis_svhn_single_label_features}
\end{figure*}

\begin{figure*}
  \centering
  \subfigure[CNN]{ 
    \includegraphics[width=0.32 \linewidth]{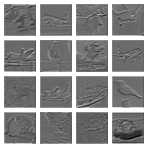} 
  } 
  \subfigure[FC CapsNet]{ 
    \includegraphics[width=0.32 \linewidth]{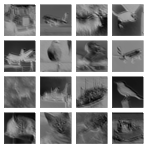} 
  }
  \subfigure[PS CapsNet]{ 
    \includegraphics[width=0.32 \linewidth]{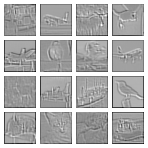} 
  }
  \caption{The visualized first convolution layer's feature map results of single-label images on CIFAR10 dataset. There are 16 single-label images, which are divided by the white line.}
  \label{vis_cifar10_single_label_conv1}
\end{figure*}

\begin{figure*}
  \centering
  \subfigure[CNN]{ 
    \includegraphics[width=0.32 \linewidth]{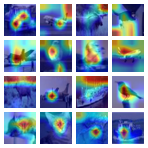} 
  } 
  \subfigure[FC CapsNet]{ 
    \includegraphics[width=0.32 \linewidth]{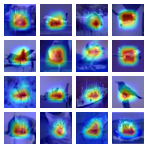} 
  }
  \subfigure[PS CapsNet]{ 
    \includegraphics[width=0.32 \linewidth]{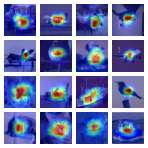} 
  }
  \caption{The visualized last layer's feature map results of single-label images on CIFAR10 dataset. There are 16 single-label images, which are divided by the white line.}
  \label{vis_cifar10_single_label_features}
\end{figure*}

\begin{figure*}
  \centering
  \subfigure[CNN]{ 
    \includegraphics[width=0.32 \linewidth]{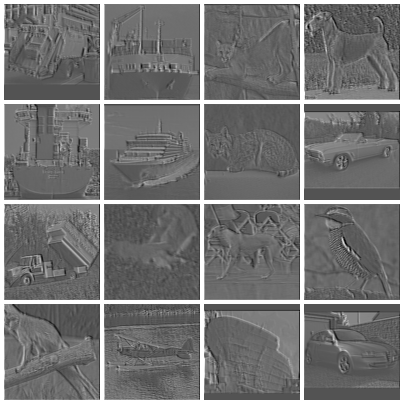} 
  } 
  \subfigure[FC CapsNet]{ 
    \includegraphics[width=0.32 \linewidth]{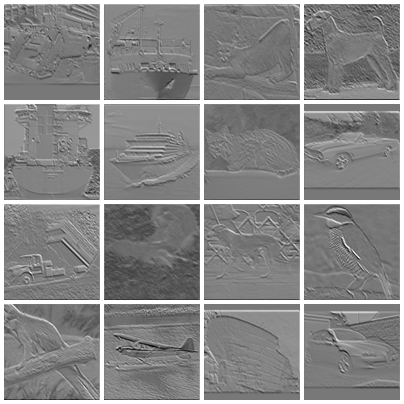} 
  }
  \subfigure[PS CapsNet]{ 
    \includegraphics[width=0.32 \linewidth]{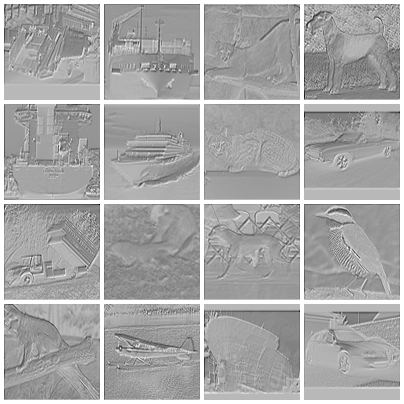} 
  }
  \caption{The visualized first convolution layer's feature map results of single-label images on STL10 dataset. There are 16 single-label images, which are divided by the white line.}
  \label{vis_stl10_single_label_conv1}
\end{figure*}

\begin{figure*}
  \centering
  \subfigure[CNN]{ 
    \includegraphics[width=0.32 \linewidth]{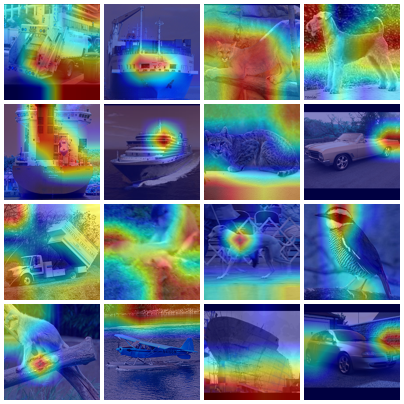} 
  } 
  \subfigure[FC CapsNet]{ 
    \includegraphics[width=0.32 \linewidth]{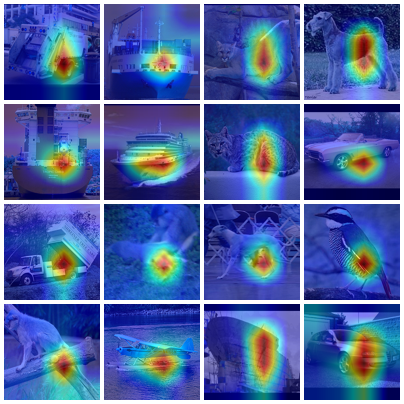} 
  }
  \subfigure[PS CapsNet]{ 
    \includegraphics[width=0.32 \linewidth]{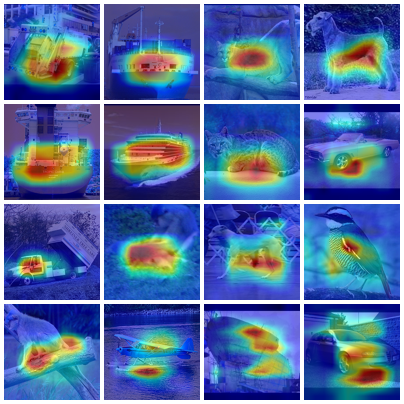} 
  }
  \caption{The visualized last layer's feature map results of single-label images on STL10 dataset. There are 16 single-label images, which are divided by the white line.}
  \label{vis_stl10_single_label_features}
\end{figure*}

\begin{figure*}
  \centering
  \subfigure[CNN]{ 
    \includegraphics[width=0.32 \linewidth]{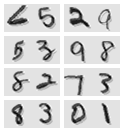} 
  } 
  \subfigure[FC CapsNet]{ 
    \includegraphics[width=0.32 \linewidth]{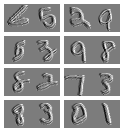} 
  }
  \subfigure[PS CapsNet]{ 
    \includegraphics[width=0.32 \linewidth]{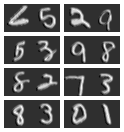} 
  }
  \caption{The visualized first convolution layer's feature map results of two-label images on MNIST dataset. There are eight two-label images, which are divided by the white line.}
  \label{vis_mnist_two_label_conv1}
\end{figure*}

\begin{figure*}
  \centering
  \subfigure[CNN]{ 
    \includegraphics[width=0.32 \linewidth]{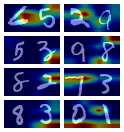} 
  } 
  \subfigure[FC CapsNet]{ 
    \includegraphics[width=0.32 \linewidth]{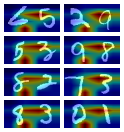} 
  }
  \subfigure[PS CapsNet]{ 
    \includegraphics[width=0.32 \linewidth]{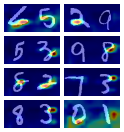} 
  }
  \caption{The visualized last layer's feature map results of two-label images on MNIST dataset. There are eight two-label images, which are divided by the white line.}
  \label{vis_mnist_two_label_features}
\end{figure*}

\begin{figure*}
  \centering
  \subfigure[CNN]{ 
    \includegraphics[width=0.32 \linewidth]{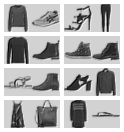} 
  } 
  \subfigure[FC CapsNet]{ 
    \includegraphics[width=0.32 \linewidth]{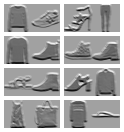} 
  }
  \subfigure[PS CapsNet]{ 
    \includegraphics[width=0.32 \linewidth]{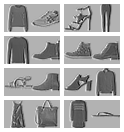} 
  }
  \caption{The visualized first convolution layer's feature map results of two-label images on FashionMNIST dataset. There are eight two-label images, which are divided by the white line.}
  \label{vis_fashionmnist_two_label_conv1}
\end{figure*}

\begin{figure*}
  \centering
  \subfigure[CNN]{ 
    \includegraphics[width=0.32 \linewidth]{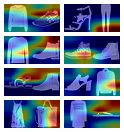} 
  } 
  \subfigure[FC CapsNet]{ 
    \includegraphics[width=0.32 \linewidth]{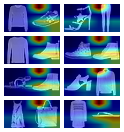} 
  }
  \subfigure[PS CapsNet]{ 
    \includegraphics[width=0.32 \linewidth]{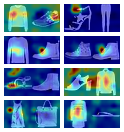} 
  }
  \caption{The visualized last layer's feature map results of two-label images on FashionMNIST dataset. There are eight two-label images, which are divided by the white line.}
  \label{vis_fashionmnist_two_label_features}
\end{figure*}

\begin{figure*}
  \centering
  \subfigure[CNN]{ 
    \includegraphics[width=0.32 \linewidth]{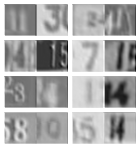} 
  } 
  \subfigure[FC CapsNet]{ 
    \includegraphics[width=0.32 \linewidth]{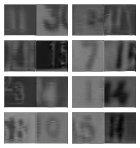} 
  }
  \subfigure[PS CapsNet]{ 
    \includegraphics[width=0.32 \linewidth]{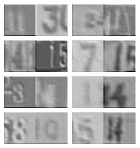} 
  }
  \caption{The visualized first convolution layer's feature map results of two-label images on SVHN dataset. There are eight two-label images, which are divided by the white line.}
  \label{vis_svhn_two_label_conv1}
\end{figure*}

\begin{figure*}
  \centering
  \subfigure[CNN]{ 
    \includegraphics[width=0.32 \linewidth]{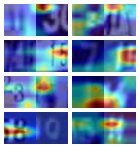} 
  } 
  \subfigure[FC CapsNet]{ 
    \includegraphics[width=0.32 \linewidth]{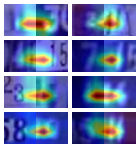} 
  }
  \subfigure[PS CapsNet]{ 
    \includegraphics[width=0.32 \linewidth]{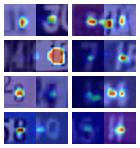} 
  }
  \caption{The visualized last layer's feature map results of two-label images on SVHN dataset. There are eight two-label images, which are divided by the white line.}
  \label{vis_svhn_two_label_features}
\end{figure*}

\begin{figure*}
  \centering
  \subfigure[CNN]{ 
    \includegraphics[width=0.32 \linewidth]{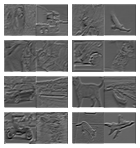} 
  } 
  \subfigure[FC CapsNet]{ 
    \includegraphics[width=0.32 \linewidth]{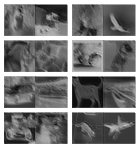} 
  }
  \subfigure[PS CapsNet]{ 
    \includegraphics[width=0.32 \linewidth]{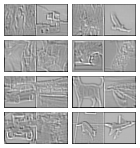} 
  }
  \caption{The visualized first convolution layer's feature map results of two-label images on CIFAR10 dataset. There are eight two-label images, which are divided by the white line.}
  \label{vis_cifar10_two_label_conv1}
\end{figure*}

\begin{figure*}
  \centering
  \subfigure[CNN]{ 
    \includegraphics[width=0.32 \linewidth]{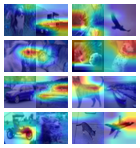} 
  } 
  \subfigure[FC CapsNet]{ 
    \includegraphics[width=0.32 \linewidth]{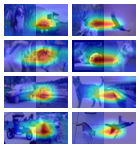} 
  }
  \subfigure[PS CapsNet]{ 
    \includegraphics[width=0.32 \linewidth]{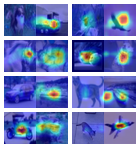} 
  }
  \caption{The visualized last layer's feature map results of two-label images on CIFAR10 dataset. There are eight two-label images, which are divided by the white line.}
  \label{vis_cifar10_two_label_features}
\end{figure*}

\end{document}